# A Dual Power Grid Cascading Failure Model for the Vulnerability Analysis

Tianxin Zhou*, Xiang Li*, Haibing Lu†
*Department of Computer Science & Engineering,
†Dept. of Information Systems & Analytics
Santa Clara University
Santa Clara, USA
{tzhou,xli8,hlu}@scu.edu

***Abstract*—Considering the attacks against the power grid, one of the most effective approaches could be the attack to the transmission lines that leads to large cascading failures. Hence, the problem of locating the most critical or vulnerable transmission lines for a Power Grid Cascading Failure (PGCF) has drawn much attention from the research society. There exists many deterministic solutions and stochastic approximation algorithms aiming to analyze the power grid vulnerability. However, it has been challenging to reveal the correlations between the transmission lines to identify the critical ones. In this paper, we propose a novel approach of learning such correlations via attention mechanism inspired by the Transformer based models that were initially designated to learn the correlation of words in sentences. Multiple modifications and adjustments are proposed to support the attention mechanism producing an informative correlation matrix, the Attention Matrix. With the Attention Ranking algorithm, we are able to identify the most critical lines. The proposed Dual PGCF model provide a novel and effective analysis to improve the power grid resilience against cascading failure, which is proved by extensive experiment results.**

## I. Introduction

With emerging smart grid technologies, the level of potential threats has been escalating in the cyber space [1]. The threats to the transmission line failures, in particular, can be further aggravated by the cascading failure effect which leads to large scale blackouts. Based on historical data [2], cascading failure is almost always the top one contributor to the massive blackouts in power grids. Hence, it is crucial to establish an effective vulnerability analysis to mitigate the cascading effect by identifying the most critical lines of the power grid.

Since the cascading effect evolves in a sequence of events, we could generalized the PGCF process into sets of failures as in generations which we will give a formal definition in Sec. III-B. There are two types of transmission lines in every generation of the PGCF: the critical transmission lines that their own failure will trigger other transmission lines to fail, the *Initiatives*, and the vulnerable transmission lines that fails because of others' failure, the *Passives*.

Ideally, if the state of all transmission lines can be calculated when a set of *Initiatives* failed initially, we could foresee the failing *Passives* and activate a real-time protection to prevent the cascading failure. However, the protection operations, such as generation redispatch or load shedding, are expensive. In 2021, $16,561$ GWh of generation re-dispatch approximately costed German €221 million [3]. The cost of load shedding is significantly high even after the minimization in [4] and it is considered "much more expensive than generation redispatch" [5].

A more reasonable approach is to protect *Initiatives* and *Passives* in advance. If we could locate the top *Initiatives/Passives*, they can be replaced with tougher materials or upgraded to higher capacities. When the *Initiatives* are protected, the chain effect of the cascading failure is interrupted and the blackout is restrained. When the *Passives* are protected, a direct control of the PGCF scale is enforced and the damage is mitigated. Protecting both *Initiatives* and *Passives* could proactively mitigate and even revent PGCF.

The methods for PGCF vulnerability analysis can be divided into two general directions. First direction is focusing on the deterministic solution which relies on the accurate calculation of the power flow for the entire power grid [6]–[9], but it has a scalability issue because the power flow computation is a $N-k$ problem which is considered intractable and impractical [6], [7]. Second direction is focusing on the stochastic approach which estimates the criticality of the transmission lines by their failing probabilities [10]–[12], but the reliability of the probability distribution is questionable and it may not accurately depict the PGCF. Neither of the existing methods can accurately reveal the correlation between transmission lines, which is crucial for the vulnerability analysis and critical line identification.

Thus, rather than finding an efficient solution for the $N-k$ problem or reliable probabilities for the accurate estimation, we have devoted our efforts to the methods for the correlation revelation. In [13], we've learned that the Transformer model can be applied to simulate the PGCF. It predicts the PGCF scale and the frequently failed transmission lines close to the ground truth. However, it does not locate the critical *Initiatives* accurately enough. More importantly, the attention mechanism of the Transformer based models have the ability to reveal the correlation between the elements which [13] failed to exploit.

In this paper, we introduce a Dual PGCF model that focuses on the essence of the PGCF problem, the correlation between the transmission lines. The innovation includes the following aspects: 1) we adopted an encoder-only structure which concentrates the training on the correlations; 2) On top of the initial and final states of the PGCF, we also feed the model with the intermediate states so it has richer context when learning PGCF; and 3) we applied masking techniques to emphasize the correlation learning. With the advanced model and the profound comprehension of the attention mechanism, we are able to introduce a fresh notion of the relationship between the attention and correlation of the failed transmission lines. The knowledge of the relationship facilitates the proposal of the algorithms that identifies the most critical *Initiatives* and the most vulnerable *Passives*.

The Dual PGCF model enables the perception of cascading failures from an unconventional perspective: rather than the accurate prediction of the PGCF processes, the model can expose hidden relationships among the transmission lines, which are not attainable via existing methods.

Experiment results further proved the effectiveness of the Dual PGCF model. When comparing the top *Initiatives/Passives* with the selection from three baseline algorithms via multiple metrics, the top *Initiatives/Passives* consistently outperforms the baseline algorithms.

Our contributions are summarized as follows.

- We proposed the Dual PGCF model enhanced on learning the correlation of the failed transmission lines. Masking techniques were applied to enforce the concentration on the correlation learning.
- We proposed a new PGCF format that provides the most necessary information and is a customized fit for training.
- We proposed a perspective that resolves the attention mechanism and its significance to reveal the correlation of transmission lines. We further exploited the attention to identify the critical transmission lines via the proposed algorithms.
- We conducted a series of experiments that produces evidently positive outcomes from the comparison with baseline algorithms.

**Organization.** The rest of the paper is organized as follows. Section II reviews the related works. Section III introduces the Dual PGCF model, the attention extraction and the attention ranking algorithms. Section IV provides the experiments that compares the top *Initiatives/Passives* with the selection from other baseline algorithms. We conclude the paper in Section V.

## II. Literature Review

The research of power grid vulnerability has been diversified into various perspectives. Among all, two general directions exist: the deterministic approaches [5]–[9], [14]–[18] and the stochastic approaches [10]–[13], [19]–[30]. The advantage of the deterministic approaches are the implementation of the power flow calculation which accurately simulates every cascading process with different initial failures. The accurate simulation could establish effective solutions for mitigating the PGCF after it begins. However, these approaches are considered impractical and intractable when it comes to identifying the top $k$ most vulnerable lines in a power grid with $n$ lines as it requires a calculation of $\binom{n}{k}$ possible initial failures to accomplish that [6], [7]. On the other hand, the stochastic approaches have the ability of approximately and efficiently estimating the most critical lines.

Since the most prominent disadvantage of deterministic models is the computational expense, the stochastic researches have been conducted aiming to approximate the computation with probability functions based on the power network properties and/or the graphical properties of the network, such as load, capacity, links. The probability functions could provide the weights or the rates for a Markov Chain model to simulate the transition of the network from one state to another [10], [11], [19]–[21].

Another perspective to build a stochastic model is to, first, reform the power network into a graph with nodes representing the transmission lines and edges representing the interaction between the transmission lines [23]. Secondly, historical or simulated PGCF data of the transmission line failure frequency is statistically (weighted average or similar formula) assigned to the edges to quantify the interaction between the nodes [12], [23]–[26]. With the interaction graph as the cornerstone, the expectation maximization algorithm is applied to located the key links in [24]. The interaction matrix is further extended to multiple interaction matrices that specifies the correlation of every generation during the PGCF process in [25]. [26] builds a Generative Adversarial Network to identify the uncovered interaction.

Although the interaction graph is an effective and efficient high-level approach, the probability acquirement is relatively debatable. As the Machine Learning (ML) techniques becomes the trend of acquiring more reliable probabilities, multiple studies achieved more realistic results via ML models [27]–[29], except the results are still too broad to be applicable for a more precise vulnerability analysis. Additionally, the ML models may learn the probability distribution but the transition from probabilities to the correlations is not yet discovered.

Deep Learning (DL) techniques have also been widely implemented to tackle different power grid tasks [31]. However, limited research was conducted with a Deep Neural Network (DNN) focusing specifically on the PGCF [13], [26], [30]. Both [30] and [26] implemented a more complex neural network than the aforementioned ML approaches. They may have improved the vulnerability analysis by classifying the lines based on the level of PGCF scale they could cause. The broad definition of the levels are yet to be meaningful for a more practical analysis. In addition, the essential correlation of the transmission lines remain hidden. The transformer model implementation in [13] did focus on learning the correlation and it proved that the transformer model can be applied to the PGCF problem. However, the model may not be the best fit for vulnerability analysis because it was focusing on the cascading failure prediction but overlooked the learning of the correlations. The Revelation of the correlations opens the opportunities to develop a more reliable and substantial vulnerability analysis.

## III. The Dual PGCF model

As the transformer based models have been proven to be effective on learning the context for many different tasks with its attention mechanism, we first attempted on training a Transformer model with the PGCF samples [13]. However, there are many technical issues with the approach.

The foremost issue is that the data is unbalanced, because there are much less initial failures than end failures. The unbalanced data could introduce a certain level of difficulty for the model to find focus. A lot of noises remain even after converging. The next issue is that, when representing transmission lines by their ID, the order of the failed lines in the input sequence could affect the training process. However, within each generation of the PGCF, it is indifferent if one transmission line failed before the other transmission line. Hence, the model may learn irrelevant information.

To construct a new vulnerability analysis method that effectively reveals the correlations without the issues in [13], we build a Dual PGCF model expecting to accomplish the mission in a practical and adequately reliable way.

The construction of the unprecedented vulnerability analysis aims to resolve a series of problems: 1) build the model emphasizing on the context learning; 2) transform the PGCF samples into the trainable data while retaining most of the context; 3) interpret the context; 4) refine the context to be concentrated on the failed transmission lines; and finally, 5) quantify the context to complete the vulnerability analysis.

### A. Model Structure

The original Transformer model was designed to accomplish the language translation task with its encoder-decoder structure that learns two different vocabularies. The PGCF, however, contains only one "vocabulary", the transmission lines. Therefore a structure with only the encoder is sufficient to be the backbone of the Dual PGCF model.

The encoder only structure simplifies the learning process and allows our model to focus on the correlation between the transmission lines. It is crucial because we need to learn which transmission lines are "attracting more attention" from others' failures or "paying more attention" to others' failures. Learning the context or "attention" is more meaningful for the vulnerability analysis as we have discussed in Section I.

According to the study by Xiong *et al.* [32], the placement of the layer normalization affects the model convergence. We exploit their findings by placing the layer normalization before the multi-head attention layer in order to achieve a smoother convergence although such training improvement won't affect the results significantly.

### B. PGCF Transformation

The goal of PGCF Transformation is to transform a raw PGCF sample to a sequence of vectors representing the failed line—a mathematical form that the model can understand better. We begin the transformation process by the definition of the raw PGCF sample with Eq. (1). Starting from the $1_{st}$ generation, the set of the initial failures, there are $G$ generations with each generation $L_t$ is a set of $n_t$ failures at generation $t$. The ID of the failed transmission line is denoted as $l_{(i,t)}$ with $i \in (1, 2, ..., n_t)$.

$$\begin{aligned} L_t &= \{l_{(1,t)}, l_{(2,t)}, ..., l_{(n_t,t)}\} \\ PGCF &= \{L_t\}_{t=1}^{G}. \end{aligned} \tag{1}$$

In [13], $L_1$ and $L_G$ are used as the input and target which leads to an unbalanced data issue as $|L_1|$ could be much smaller than $|L_G|$. It is also not the best format to convey as many context as possible for training. The raw PGCF sample format disregards two factors: the independence between the failed transmission lines in one generation and the progress of all generations. The independence of the failures in one generation relates to the fact that the failure of $L_t$ only depends on the failure of $L_{t-1}$, and, the order of all $l_{(i,t)}$ in $L_t$ is irrelevant. Hence, we have to avoid the model to be affected by the inconsistent order of the inputs or it could be a disturbance during the training. In addition, $L_t$ contains solely the transmission line IDs, meaning that the model won't be able to receive any information regarding the failed generation.

Therefore, instead of using the IDs of the failed transmission lines as the input and the target, we transformed each PGCF sample into $G$ sequences and each sequence contains $N$ elements. The index of the elements represents the ID and the elements themselves indicates the failed generation. In consequence, one generation $t$ ($t = 1, 2, ..., G$) of a PGCF process (Eq. (1)) can be transformed to:

$$\boldsymbol{gen_t} = \{g_i : g_i \in \{1, 2, ..., t\}\}_{i=1}^{N} \tag{2}$$

Each element $g_i$ equals to the failed generation of the transmission line $i$ or 0 if it did not fail. For instance, $\boldsymbol{gen_1}$ is a sequence of 1s and 0s representing initial failures and the healthy lines in the $1_{st}$ generation. Sequence $\boldsymbol{gen_G}$ represents the failed generation of all the lines or 0 for those did not fail.

Since one PGCF sample could produce $G$ vectors, we could have $G-1$ pairs of the inputs and the targets. An example of $\boldsymbol{gen_t}$ as the input and $\boldsymbol{gen_{t+1}}$ as the target at generation $t$ would be:

$$\begin{cases} \boldsymbol{gen_{inp}} = \{g_i : g_i \in \{0, 1, 2, ..., t\}\}_{i=1}^{N} \\ \boldsymbol{gen_{tar}} = \{g'_i : g'_i \in \{g_i, t+1\}\}_{i=1}^{N} \end{cases}. \tag{3}$$

If $g_i$ is not 0, $g_i$ shall be the same as $g'_i$. If $g_i$ is 0, $g'_i$ may be 0 or $t+1$.

We call this transformation a dual representation because it contains both the information of transmission line ID and the generation in which it failed. That is also why we named our model the dual model. With this dual representation, our model is capable of learning not only the correlation consistently but also the information sufficiently.

The dual representation is further vectorized via two embedding matrices. Because the indices of the elements in $\boldsymbol{gen_t}$ could also be perceived as the positions of the transmission lines in the real power grid, we call the first embedding matrix the position embedding matrix. The other is called the generation embedding matrix which vectorizes the generation information. Fig. 1 summarizes the vectorization process. We obtain the vector representation of $\boldsymbol{gen_t}$ by the concatenation of the position vector and the generation vector. The vector representation includes the information the model requires and is in the format that the model understands.

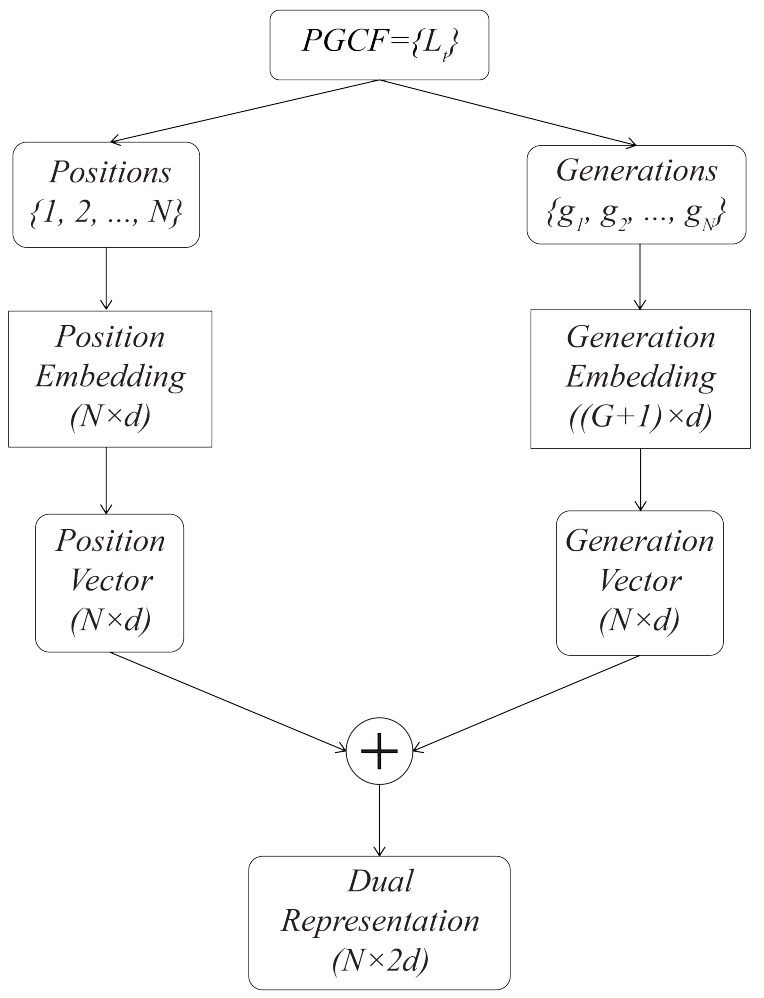

Fig. 1: Dual Representation to Vector Representation

### C. Attention Mechanism

The advantage of the Transformer based model is its attention mechanism that learns correlations between elements. However, the interpretation of the attention between the failed transmission lines in PGCF is not straightforward.

The attention equation is defined as

$$att(Q, K, V) = softmax(\frac{Q \cdot K^T}{\sqrt{d_k}}) \cdot V. \quad (4)$$

Although the matrices $Q, K, V$ have different names, they are actually the same matrices with the dimension of $n \times d$ which will be the vector representation of the input. $\sqrt{d_k}$ is a constant value that helps the convergence [33]. The $softmax$ function will output an attention weight matrix $AW_{(N,N)}$ (Eq. (5)) that can be interpreted as a probability matrix. Because the $softmax$ function is normalizing the last dimension, each row of the attention weight matrix will be summed up to 1. In [33], the attention vectors are the multiplications of the attention weight matrix and the $Value$ matrix. In our study, however, we consider the the attention weights as the attention, since we seek the correlation more than the final prediction and the attention weight matrix is the description of the correlations.

When we trying to identify the most critical *Initiatives*, we focus on their ability of initiating the failure of the others. The most critical *Initiatives* could lead to most of the other transmission lines to fail but themselves may not fail as easy due to others' failure. Conversely, regarding the most vulnerable *Passives*, even though they have a higher possibility to fail, their failure may not necessarily be the reason for the failure of the other transmission lines. With the correlation description by $AW$, we are able to quantify the criticality and the vulnerability of the transmission lines.

In particular, $a_{i,j}$ in $AW$ represents the attention paid to transmission line $j$ from transmission line $i$, or the likelihood of $j$'s failing is a result of $i$'s failure. Furthermore, each row $i$ of $AW$ is a vector represents the attention of transmission line $i$ drew from every other transmission lines. Also, each column $j$ represents the attention of transmission line $j$ paid to every other transmission lines. In another word, the probability distribution of transmission line $i$ being passively deactivated by the failure of others is represented by the row $i$. And, the probability distribution of transmission line $j$'s failure actively initiate the failure of others is represented by the column $j$.

$$AW_{(N,N)} = \begin{pmatrix} a_{1,1} & a_{1,2} & \cdots & a_{1,N} \\ a_{2,1} & a_{2,2} & \cdots & a_{2,N} \\ \vdots & \vdots & \ddots & \vdots \\ a_{N,1} & a_{N,2} & \cdots & a_{N,N} \end{pmatrix} \quad (5)$$

### D. Masking

Because we could train the model by pairs of two consecutive generations (Eq. (3)), $|\{g'_i : g'_i \neq g_i\}|$ is not going to be significantly large, but the number of 0s in any $\boldsymbol{g_t}$ could be a lot larger than those are not, especially for the earlier generations. To remove this imbalance between zeros and non-zeros and have the model focus on learning the failed transmission lines rather than those that are alive, we decided to employ a particular masking scheme before the forward propagation and the backward propagation.

First, we mask the alive transmission lines in the input so the attention mechanism (Section III-C) won't calculate the attention from them. Second, we mask the loss for all the transmission lines except those failed at generation $t + 1$. Therefore, during the backward propagation, only the attention of the transmission lines failed in generation $t + 1$ will be updated.

For example, if we have a sample $\boldsymbol{g}$, the input mask will be a sequence of

$$inpM = \left\{ inpM_i \middle| \begin{matrix} 0, & g_i = 0 \\ 1, & g_i > 0 \end{matrix} \right\}_{i=1}^{N},$$

and the target mask will be

$$tarM = \left\{ tarM_i \middle| \begin{matrix} 0, & g'_i < t+1 \\ 1, & g'_i = t+1 \end{matrix} \right\}_{i=1}^{N}.$$

Once we broadcast $inpM$ into an $N \times N$ matrix, the attention calculation can be masked to focus only on the transmission lines failed in generation 1 to $t$ (Eq. (6)).

$$att(Q, K, V) = (softmax(\frac{Q \cdot K^T}{\sqrt{d_k}}) \times inpM) \cdot V. \quad (6)$$

The target mask will be applied to the cross entropy loss function before the backward propagation. The total loss $L$ is a summation of the loss $L_i$ for each of the transmission line $i$, which is also a summation after an element-wise matrix multiplication by the label matrix $Y$ and the probability matrix $P$ (Eq. (7)). Each row of $Y$ will be an one-hot encoding with $y_{(i,g)} = 0$ if transmission line $i$ did not fail at generation $g$, otherwise $y_{(i,g)} = 1$. And, for each row of $P$, $P_{(i,g)}$ is the probability distribution of transmission line $i$ failing at generation $g$.

$$
\begin{aligned}
L &= -\sum_{i=1}^{N}\sum_{g=0}^{t+1} Y_{(i,g)} \times \log P_{(i,g)} \\
&= -\sum_{i=1}^{N}\{L_i | L_i = \sum_{g=0}^{t+1} Y_{(i,g)} \times \log P_{(i,g)}\}
\end{aligned} \tag{7}
$$

If we apply a mask $tarM$ to $L_i$ in Eq. (7), we have

$$
L = -\sum_{i=1}^{N}\{L_i \times tarM_i\}. \tag{8}
$$

When the model tries to minimize the loss with Eq. (6), only the loss for the transmission lines failed in $t+1$ will be updated because $tarM_i$ for others will be 0.

With Eq. (6) and Eq. (8), our model could be trained to focus only on the correlation between the transmission lines failed in generation 0 to $t+1$. In the end, we expect the attention matrix $AW$ to possess exactly the information we need without the noise from the transmission lines that are not contributing to the PGCF.

**Algorithm 1** Attention Extraction

**Input:** Pre-trained model $P$, Sample set $S$, Number of transmission lines $N$

**Output:** Initiative correlation matrix $ICM(N \times N)$, Passive correlation matrix $PCM(N \times N)$

*Initialisation*: $ICM \leftarrow 0_{N\times N}$, $PCM \leftarrow 0_{N\times N}$

1: **for** Input $inp$ and Target $tar$ in $S$ **do**
2: $predictions, AW_{(h,N,N)} \leftarrow P(inp)$
3: $AW_{(N,N)} \leftarrow \|AW_{(h,N,N)}\|_2$
4: $index \leftarrow$ *where* $tar_i$ *is* $max(tar)$
5: $scale \leftarrow \{scale_i | scale_i = inp_i / max(inp)\}_{i=1}^{N}$
6: $PCM[index] \leftarrow PCM[index] + AW[index] \times scale$
7: $index \leftarrow$ *where* $inp_i$ *is* 1
8: $mask \leftarrow \left\{ mask_i \middle| \begin{matrix} 0, & tar_i \neq max(tar) \\ 1, & tar_i = max(tar) \end{matrix} \right\}_{i=1}^{N}$
9: $ICM[:, index] \leftarrow ICM[:, index] + AW[:, index] \times mask$
10: **end for**
11: **return** $ICM$, $PCM$

## E. Attention Analysis

Once the Dual PGCF model is trained with the input/target in Dual representation and the masking scheme application, it has the ability to present the attention between the failed transmission lines in any input/target pairs. We know the attention can be related to the criticality/vulnerability of the transmission lines but we still need a method to quantify the attention and complete the final step of the vulnerability analysis.

*1) Attention Extraction:* Due to the multi-layer structure of the Dual PGCF model, we have $M$ ($M$ is the number of the encoder layers) attention weight matrices for each sample. Because the last encoder layer is the closest to produce the softmax probability for the final prediction, the attention weights from the last layer could be much more significant and precise to reveal the correlation between the transmission lines. Thus, we only use attention weights from the last layer for attention extraction.

In [33], the authors found it beneficial to not calculate the attentions all at once for the $d$ dimensions of $Q$, $K$, $V$ (Eq. (4)). In practice, they divide $d$ into $h$ pieces for an additional "head" dimension. After the attention calculation, the $(n, d/h)$ attentions will be concatenated back into one for the following layers. That means, we have $h$ $AW$s to conclude to one attention matrix. In order to maintain most of the information, we project the "head" dimension on to the last two dimensions of the $AW$ by the L2 normalization and the head dimension will be reduced.

However, one single $AW$ from one sample could not carry the complete correlation between all transmission lines. We still have a great number of $AW$s from the entire sample set $S$. Because each sample has different initial failures, it could only produce correlations between a limited number of the transmission lines. Also, because of the opposite interpretation of the transmission lines as in *Initiatives* or *Passives*, we need an algorithm to help us collect the correlations with these two different perspectives.

To extract the attentions for *Passives*, we want to collect the rows that represent the transmission lines failed in the last generation in the target (step 4 in Alg. 1). However, the *Initiatives* in different generation should contribute to the failure of the *Passives* differently. The newly failed *Initiatives* should have higher weights .

Hence, at step 5 and 6 in Alg. 1, we accumulate the attention of *Passives* failed at the last generation weighted the failed generation of the input *Initiatives*. The earlier the *Initiatives* fails, the less attention to be accumulated.

To extract the attentions for the *Initiatives*, we decided to collect the columns that represent the *Initiatives* failed in the initial generation (step 7 in Alg. 1). Because we applied $tarM$ to the loss function during the training process (Eq. (8)), only the rows representing the *Passives* failed in the last generation will be updated. Hence, step 8 creates a mask to ensure only the attention paid to the *Passives* failed in the last generation will be accumulated to the *Initiatives* failed in the initial generation (step 9 in Alg. 1).

*2) Attention Ranking:* The last step to complete the vulnerability analysis is to find the top *Initiatives/Passives* based on the correlation matrices. If we consider the correlation matrices as probability distributions, the expectations of the *Initiatives/Passives* can be defined as

$$
\begin{aligned}
Expectation_P_i &= \sum_{k=1}^{N} PCM_{(i,k)} \\
Expectation_I_j &= \sum_{k=1}^{N} ICM_{(k,j)}.
\end{aligned} \tag{9}
$$

Because the failure of a *Passive* is independent from the failure of any other *Passives*, the rank of the most vulnerable *Passives* could simply be the rank of all the $Expectation_P$s.

However, when ranking the *Initiatives*, we have to consider the different contribution to the PGCF from all the initial failures. Some of the *Initiatives* may contribute more than others if failed initially. And, some *Initiatives* may leads to larger PGCF only if they fail along with other *Initiatives*.

We could choose the *Initiative* $j_1$ with the highest $Expectation_R$ to begin with, but the choice of the second *Initiative* $j_2$ must consider the maximization of $Expectation_R_{(j_1,j_2)}$ but not a simple summation:

$$Expectation_R_{(j_1,j_2)} = \sum_{k=1}^{N} max(ICM_{(k,j_1)}, ICM_{(k,j_2)}). \tag{10}$$

The choice of $j_2$ should be a complement to $j_1$. For the transmission lines that $j_1$ is not paying much attention to, $j_2$ should pay the most attention comparing to others . If we select $j_2$ simply by the next highest $Expectation_R$ or the sum of the $Expectation_R$ of $j_1$ and $j_2$, it may create a situation that both $j_1$ and $j_2$ are paying high attention to the same transmission lines, and the PGCF might not be maximized.

In the Attention Ranking algorithm (Alg. 2), *buffer* is updated every time a new *Initiative* is selected. It stores the maximized $Expectation_R$ to find the next *Initiative* that is expected to trigger more *Passives* to fail. At the termination of the outer loop, it is possible that there are remaining transmission lines that could not pay more attention to other transmission lines than the *Initiatives* currently selected. We could rank the rest of the transmission lines by the sum of the columns of *ICM*.

**Algorithm 2** Rank Attention

**Input:** Initiative correlation matrix $ICM(N \times N)$, Passive correlation matrix $PCM(N \times N)$
**Output:** Rank of $Initiatives/Passives$
*Initialisation*: Assign empty lists to $Initiatives$ and $Passives$.

$Passives \leftarrow$ indexes sort of the sum of *PCM* by the rows
$rank \leftarrow$ indexes sort of the sum of *ICM* by the columns
*buffer* $= ICM[:, rank[0]]$
**while** $Initiatives.size < N$ **do**
  $Expectation_R \leftarrow 0$
  $next \leftarrow 0$
  $i \leftarrow 1$
  **for** $j = i$ to $N$ **do**
    **if** $j \in Initiatives$ **then**
      continue
    **end if**
    $tempE = \sum_{k=1}^{N} max(buffer_k, ICM_{k,j})$
    **if** $tempE > Expectation_R$ **then**
      $Expectation_R = tempE$
      $next = j$
      $tempI = \{k | where\ ICM_{k,j} > buffer_k\}$
    **end if**
  **end for**
  **if** $next \geq 1$ **then**
    $Initiatives.append(next)$
    $buffer[tempI] = ICM[tempI, next]$
    increment $i$
  **else**
    break
  **end if**
**end while**
**if** $Initiatives.size < N$ **then**
  fill $Initiatives$ with the rest by the $rank$
**end if**
**return** $Initiatives, Passives$

## IV. Experiments

We design the experiments with following four stages to prove effectiveness of the Dual PGCF model on power grid vulnerability analysis: 1) introduce three baseline algorithms that evaluate the power grid vulnerability based on graph properties and power flow variation; 2) extract the feature from the trained Dual PGCF model, and obtain the top *Initiatives/Passives* with the algorithms proposed in Sec. III-E; 3) project the vulnerabilities onto the power network; 4) examine the top *Initiatives/Passives* by comparing them with the selections from baseline algorithms.

### A. Baseline Algorithms

The baseline algorithms we chose to compare with are:

- Edge Betweenness Centrality (BC)
- Edge Current Flow Betweenness Centrality (CFBC) [34]
- Line Outage Distribution Factors (LODF) [35]

Both BC and CFBC measure the importance of the transmission lines by calculating the number of possible paths through the transmission lines between nodes. The BC of a transmission line $i$ is higher than others if more shortest paths between different pairs of nodes pass the transmission line $i$. Different than the BC, the CFBC calculates the power flow through a transmission line between all the nodes. A transmission line has higher CFBC if more power flow through it from different nodes. The transmission lines with high BC and CFBC are easier to be targeted because their failure will break the network into isolated components.

Differently, LODF measures the vulnerability of a transmission line by calculating the variation of the power flow after other transmission lines' failure. For instance, assuming transmission line $B$ fails, extra power could flow through transmission line $A$. The ratio of the extra power to $A$'s original power is the LODF between $A$ and $B$. If the LODFs between $A$ and all other transmission lines are high, we consider $A$ to be vulnerable.

If we rank the transmission lines by the vulnerability calculated by the baseline algorithms, we could compare them with the rank of *Initiatives/Passives* as the proof of attention analysis being more effective in critical line identification for PGCF.

### B. Training the Dual PGCF model

*1) Data Generation:* The power flow model from [6] is used to generate the original PGCF samples. First, 2 to 8

| Power Network | Buses | Transmission Lines | Samples | Ave. Generation | Ave. Scale |
|---|---|---|---|---|---|
| PT | 375 | 203 | 100,000 | 4.0 | 8.9% |
| DE | 942 | 433 | 100,000 | 7.8 | 32.4% |
| FR | 1671 | 860 | 100,000 | 9.4 | 49.0% |

TABLE I: Data Information

transmission lines are randomly selected as initial failures. Then, iteratively, the power distribution is rebalanced and power flow equation is resolved. The lines with load higher than capacity will be tripped and marked as failed.

To prove the experiments can produce consistent results without the bias from various design of the power grid, we use the open source power grid model from PyPSA-Eur [36] to simulate three power networks based on the real power grids in Portugal, Germany, and France. The power grids are different in both design and scale. The German power network is about twice as large as the Portugal's and the French power network is about twice as large as the Germany's.

Besides the difference in design and scale, another factor that might introduce a bias to the experiments is the power consumption variation during different time periods of the year. The variation of the power consumption leads to different initial states of the power grid. Our solution to reduce this potential bias is to average the power consumption by months of 2013.

100,000 data samples over the year 2013 are generated for each power network (Tab. I). To have a general picture of the samples, we averaged the scale (number of total failures divided by total number of lines $N$) by month and collected the PGCF frequency distribution (number of PGCFs of the month divided by the total number of samples) in Fig. 2.

We transformed the samples into Dual representations via the method discussed in Section III-B. 60% of the samples will be used for training, 20% for the validation, and 20% for testing the model after training and the comparison with the baseline algorithms.

*2) Model parameters:* We initialize our Dual PGCF Model with 6 encoder layers, 8 attention heads, 256 as the dimensionality of the embedding layers and 1024 as the dimensionality of the FeedForward layers, and 32 as the batch size. The model is trained on a workstation with a CPU with 32 cores and a NVIDIA GeForce RTX 3090 GPU card. After 20 epochs of training with each power network's samples, we obtained three trained Dual PGCF models for the following experiments.

*3) F-score metrics:* Assuming the transmission line $i$ failed at generation $g_i'$ in the target set and the prediction is $\hat{g}_i$, a True Positive prediction is when $g_i'$ is the same as $\hat{g}_i$, a False Positive prediction is when they are not the same, and a False Negative prediction is when $g_i'$ is not 0 but $\hat{g}_i$ is.

The F1 scores collected by the test set are listed:

| Power Network | F1 score |
|---|---|
| PT | 0.99 |
| DE | 0.99 |
| FR | 0.99 |

TABLE II: F1 scores

Although the F1 scores are high, they are collected without including the prediction of the transmission lines that are alive. Note that, we do not need to accurately predict if a transmission line is alive. We need an accurate prediction for the failed transmission lines only in order to obtain the most expressive attention matrices.

### *C. Vulnerability Projection*

After the models was trained and the attention extracted by Alg. 1 and Alg. 2, we obtained the top *Initiatives/Passives* for each network. We first project the vulnerability and the criticality onto the real power network for a more direct visualization of the relationship between the vulnerability and the capacity of the transmission lines, and between the criticality and the average power flow. Next, we perform a statistic analysis of the test set, by comparing the failure frequency and the PGCF scale between the top *Initiatives/Passives* and the baseline algorithms.

*1) Graphical Projection:* Fig. 3 project the top 10% most vulnerable transmission lines in color red and the capacities of lines in thickness onto the DE power network. The thicker the line is, the higher capacity it possesses. If we assume the capacity is one of the major factor of the transmission line failures, we expect to see that the line with lower capacity (thinner in the figures) is more vulnerable due to others' failure. It is clear that top *Passives* have lower capacity comparing with the selection from other baseline algorithms.

On the other hand, when examining *Initiatives*, the power flow could be the factor that has the most impact to other transmission lines' failures. In Fig. 4, top 10% of the *Initiatives* are compared with other baseline algorithms with DE power network too. The thickness of the line shows the average power flow of that transmission line during year 2013. We can observe that the top *Initiatives* are more likely distributed in the heavy load regions.

The observation of Fig. 3 and Fig. 4 can be a proof of the effectiveness of identifying the most critical/vulnerable transmission lines via the rank of *Initiatives/Passives*. The top *Passives* are more likely to fail because their capacities could be lower. The top *Initiatives* are more likely to trigger others to fail because they possess more power flow to distribute once failed. Although the proof could be arguable since a small portion of the top *Initiatives/Passives* do not possess capacity or power flow as expected, we will provide more convincing experiment results next.

*2) Failure Frequency of Passives:* If we were to examine the most vulnerable *Passives*, the failing frequency after the initial failures and the overall failing frequency could be two straightforward indicators. Particularly, we would like to focus on the failing frequency of the 2nd generation because protecting the *Passives* in an earlier generation may prevent the cascading effect more efficiently.

In Fig. 5 and Fig. 6, the failing frequency of top 10% *Passives* are significantly higher than the ranks from the baseline algorithms. Out of $20,000$ samples from the test set. Except for the top $3\%$ *Passives* of PT power network, it is obvious that the attention analysis could select the most vulnerable *Passives* effectively.

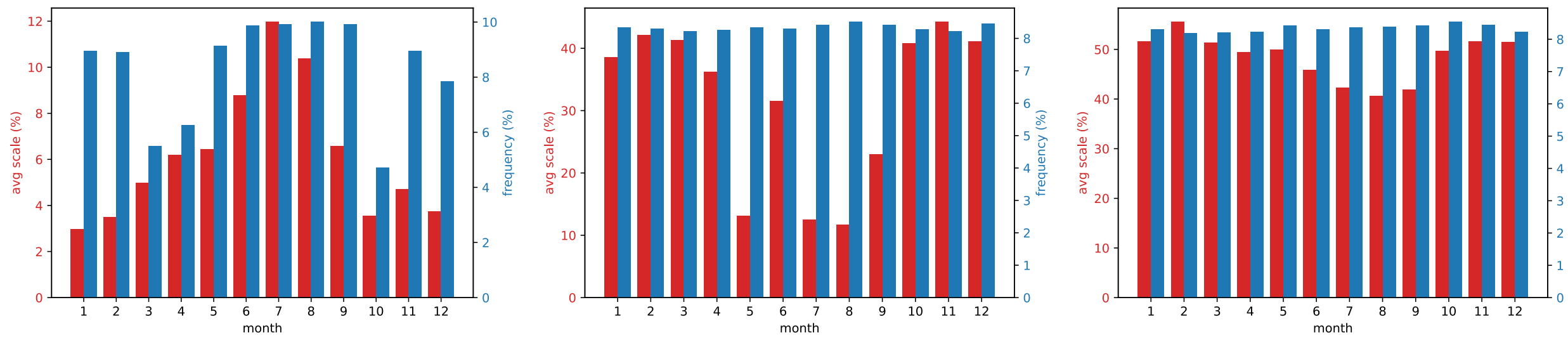

(a) PT (b) DE (c) FR

Fig. 2: Average scale (red in %) and frequency distribution (blue in %) of 100,000 samples

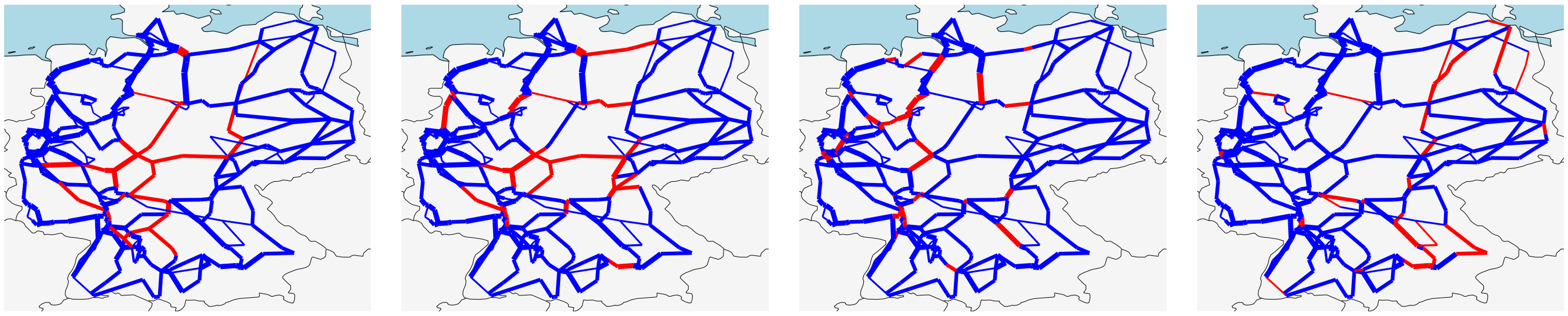

(a) BC (b) CF (c) LODF (d) Passives

Fig. 3: Top 10% of the rank (in red) vs. Capacity (in thickness) - DE

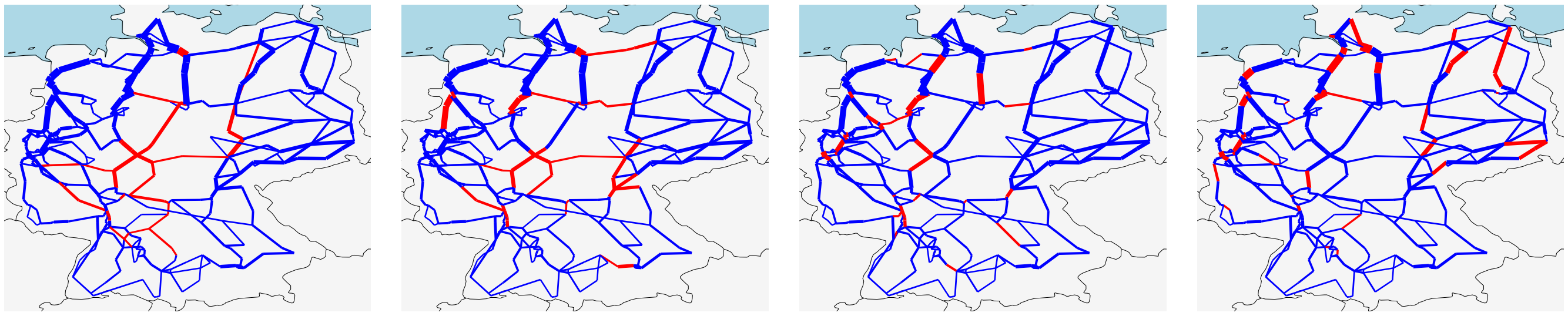

(a) BC (b) CF (c) LODF (d) Initiatives

Fig. 4: Top 10% of the rank (in red) vs. Power flow (in thickness) - DE

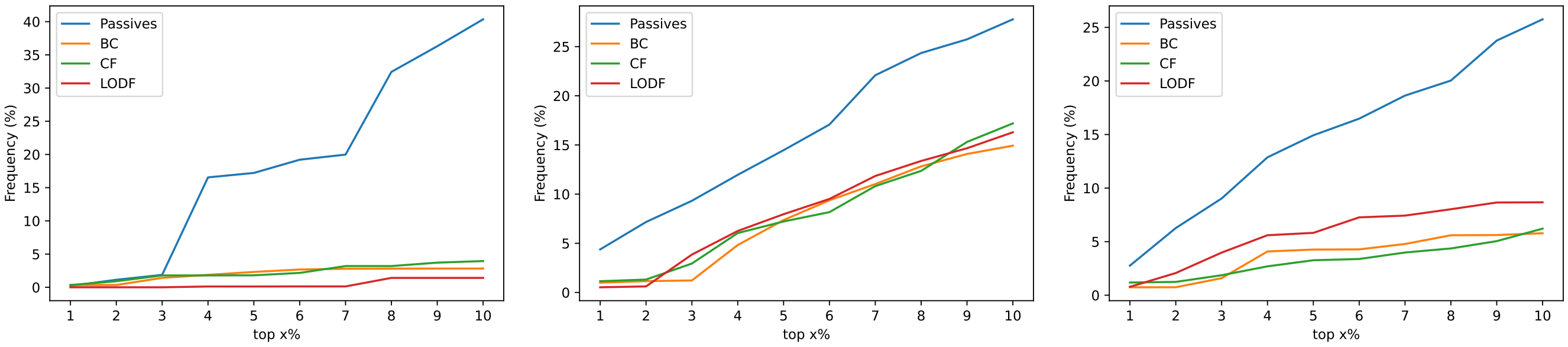

(a) Case PT (b) Case DE (c) Case FR

Fig. 5: Frequency of the top x% of the *Passives* failed in the 2nd generation

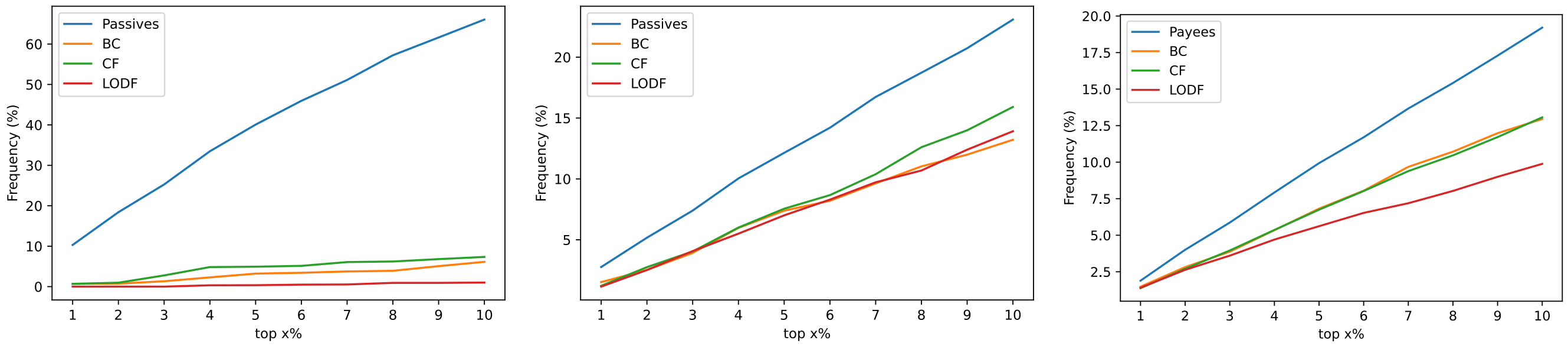

(a) Case PT (b) Case DE (c) Case FR

Fig. 6: Frequency of the top x% of the *Passives* failed in all the generations

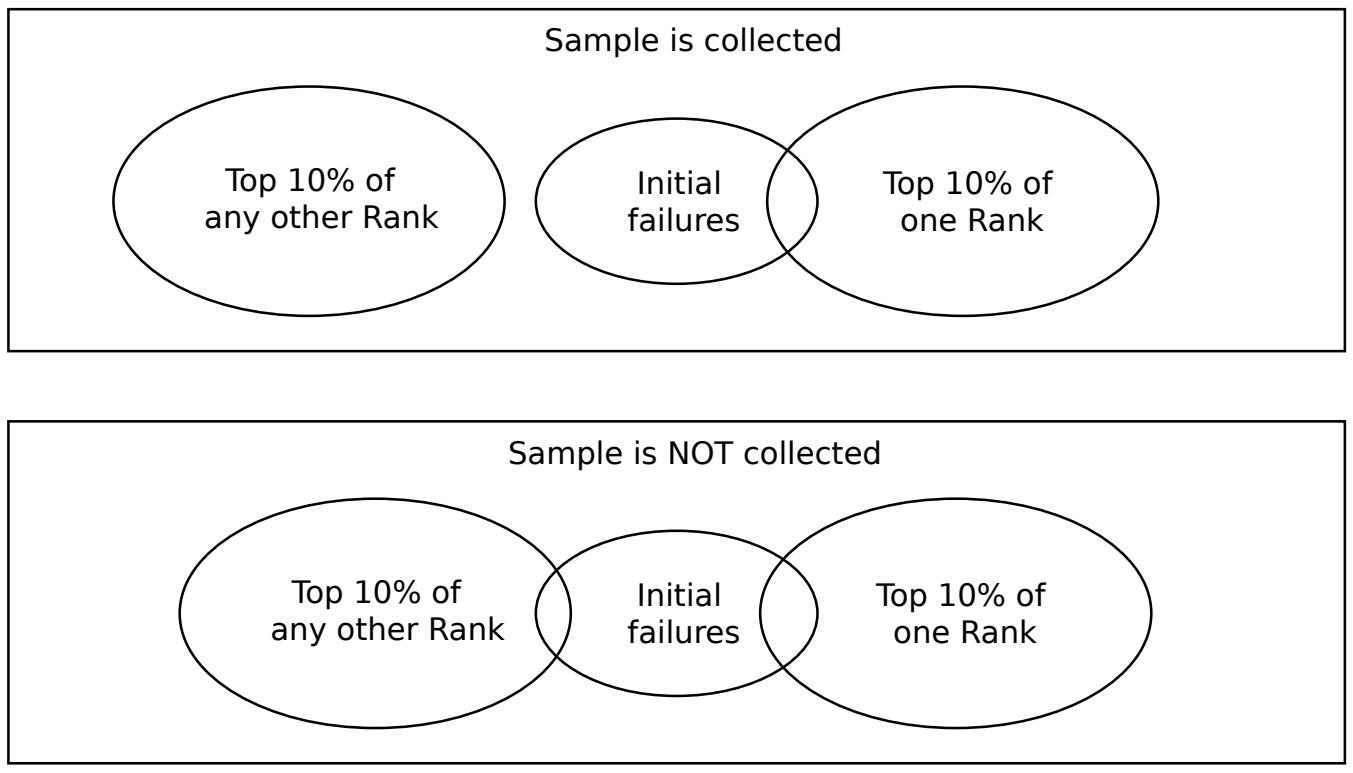

Fig. 7: Sample Selection for the PGCF Scale Experiment

*3) PGCF Scale of Initiatives:* Unlike the protection of *Passives*, protecting the most critical *Initiatives* could be more meaningful because it increases the chance of preventing the cascading effect from happening. To examine the effectiveness, if the critical *Initiatives* failed in the initial generation along with other initial failures, we could expect to see a larger average PGCF scale statistically compared to the top selection from other algorithms.

However, the sampling requires one constraint to eliminate bias. For example, if a set of initial failures include one of the most critical *Initiatives* and another line with the highest betweenness centrality, it is nearly impossible to distinguish the contribution to the PGCF scale from the two. Thus, to avoid the perplexity, we chose the samples with initial failures containing transmission lines from one and only one of the four ranks (Fig. 7).

Based on this sampling principle, the criticality of top x% selections are compared in Fig. 8 with each single bar is an average of at least $14,000$ samples outside the training set. If we compare the average PGCF scale of the samples with the initial failure contains at least one of the top x%, the top *Initiatives* significantly outperform the baselines.

There exists another observation that a descending trend appears with DE and FR power network. This behavior could be another proof that our method is effective because the lower the rank of the *Initiatives* introduce smaller PGCF scale. The PT power network, on the other hand, does not show the descending trend. Considering the PT network consists less transmission lines and behaving more volatile among other experiments (Fig. 2a and Fig. 4a), it could be that PT network is more inconsistent to be learned by a DNN. This inconsistency is also observed in [13] that the smaller networks—IEEE-118 and IEEE-300— are more difficult to learn comparing to the larger network—SciGrid.

## V. Conclusion

In this paper, we proposed a new method for the power grid vulnerability analysis that exploits the attention mechanism to reveal the correlation among the transmission lines. Different from previous deterministic approaches and other stochastic approaches, the proposed Dual PGCF model has the ability of extracting the attention and converting the attention to the correlation. It provides a practical solution to effectively and efficiently locate the most critical and vulnerable transmission lines. We prove that our method is effective via various experiments. The top *Initiatives/Passives* outperform other baseline algorithms among all the experiments.

We do believe some more sophisticated experiment designs can be developed to prove the effectiveness. We also plan to empower the model to be able to study the PGCF from other aspects such as the load level or the correlation between live transmission lines and the failed ones. And finally, we believe there are potential improvements yet to be discovered to enhance the performance of the Dual PGCF model.

## Acknowledgement

Xiang Li is supported by NSF CNS-1948550.

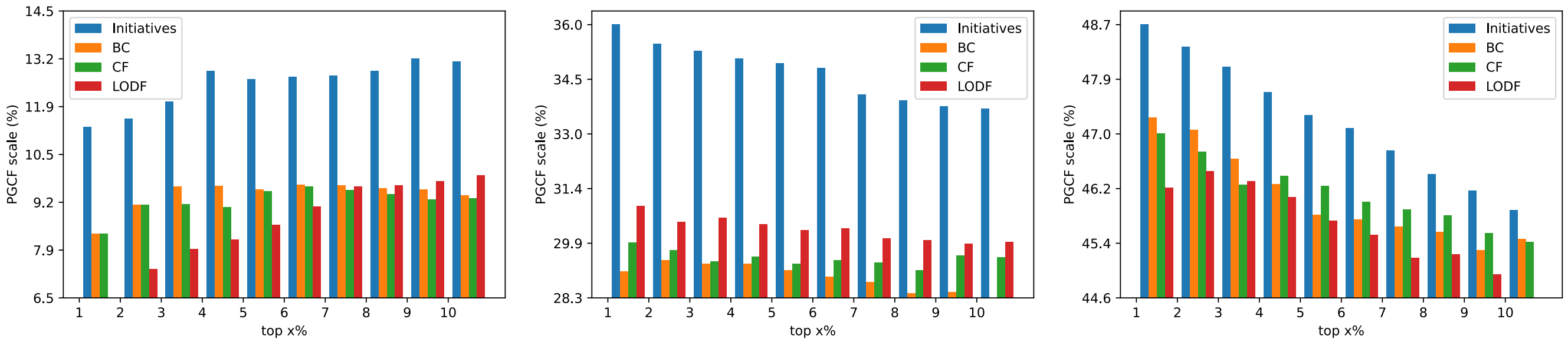

(a) Case PT (b) Case DE (c) Case FR

Fig. 8: PGCF scale with at least one initial failure is in top $x\%$ of the *Initiatives*